\pdfoutput=1
\documentclass[11pt]{article}

\usepackage[preprint]{acl}
\usepackage{times}
\usepackage{latexsym}
\usepackage{booktabs}

\PassOptionsToPackage{table}{xcolor}
\usepackage{colortbl}
\usepackage{booktabs}
\usepackage{tabularx}
\usepackage{caption}
\usepackage{geometry}
\definecolor{headergray}{RGB}{230,230,230}
\definecolor{row1}{RGB}{250,220,210}    % custom medium coral
\definecolor{row2}{RGB}{180,220,220}   % medium azure
\definecolor{row3}{RGB}{230,230,230}   % very light gray
\definecolor{row4}{RGB}{255,250,205}   % lemon chiffon
\definecolor{row5}{RGB}{224,240,230}   % light cyan

\definecolor{eval1}{RGB}{255,245,238}
\definecolor{eval2}{RGB}{240,255,255}
\definecolor{eval3}{RGB}{245,255,250}
\definecolor{eval4}{RGB}{255,250,205}
\definecolor{eval5}{RGB}{224,255,255}
\definecolor{eval6}{RGB}{250,240,230}
\definecolor{eval7}{RGB}{255,255,240}
\usepackage[T1]{fontenc}
\usepackage[utf8]{inputenc}

\usepackage{microtype}

\usepackage{inconsolata}

\usepackage{graphicx}
\usepackage{multirow}
\title{Agent Ideate: A Framework for Product Idea Generation from Patents Using Agentic AI}

\author{Gopichand Kanumolu\textsuperscript{1} \enspace \enspace Ashok Urlana\textsuperscript{1,2}  \enspace \enspace Charaka Vinayak Kumar\textsuperscript{1} \\ \enspace \enspace \textbf{Bala Mallikarjunarao Garlapati\textsuperscript{1}} \\
TCS Research, Hyderabad, India\textsuperscript{1} \enspace
IIIT Hyderabad\textsuperscript{2} \enspace \enspace \enspace \enspace \enspace \enspace \\
\texttt{\{gopichand.kanumolu, ashok.urlana, charaka.v, balamallikarjuna.g\}}@tcs.com \\ \texttt{\nolinkurl{ashok.u@research.iiit.ac.in}}
}
\begin{document}
\maketitle
\begin{abstract}
Patents contain rich technical knowledge that can inspire innovative product ideas, yet accessing and interpreting this information remains a challenge. This work explores the use of Large Language Models (LLMs) and autonomous agents to mine and generate product concepts from a given patent. In this work, we design \textbf{\textit{Agent Ideate}}, a framework for automatically generating product-based business ideas from patents. We experimented with open-source LLMs and agent-based architectures across three domains: Computer Science, Natural Language Processing, and Material Chemistry. Evaluation results show that the agentic approach consistently outperformed standalone LLMs in terms of idea quality, relevance, and novelty. These findings suggest that combining LLMs with agentic workflows can significantly enhance the innovation pipeline by unlocking the untapped potential of business idea generation from patent data.
% \begin{enumerate}
%     \item Task definition
%     \item Why it is useful
%     \item What is the methodology
%     \item What is the end results
% \end{enumerate}
\end{abstract}

\section{Introduction}
With the rapid advancement of large language models (LLMs), there is growing interest in leveraging these models for tasks such as scientific discovery and innovation support. However, generating viable and actionable product ideas from patents requires not only comprehension of complex technical content but also creativity, domain knowledge, and market awareness \cite{urlana2024llms}.  Patents are legal documents that protect inventions and promote technological innovation \cite{mossoff2000rethinking}, but their complex and technical language poses unique challenges. Despite the wealth of technical insights contained within patent documents, generating product business ideas from patents remains an underexplored area \cite{jiang2024artificial}.  
\begin{figure}[t]  
    \centering  
    \includegraphics[clip, trim=4.5cm 3cm 10cm 2cm,width=\columnwidth]{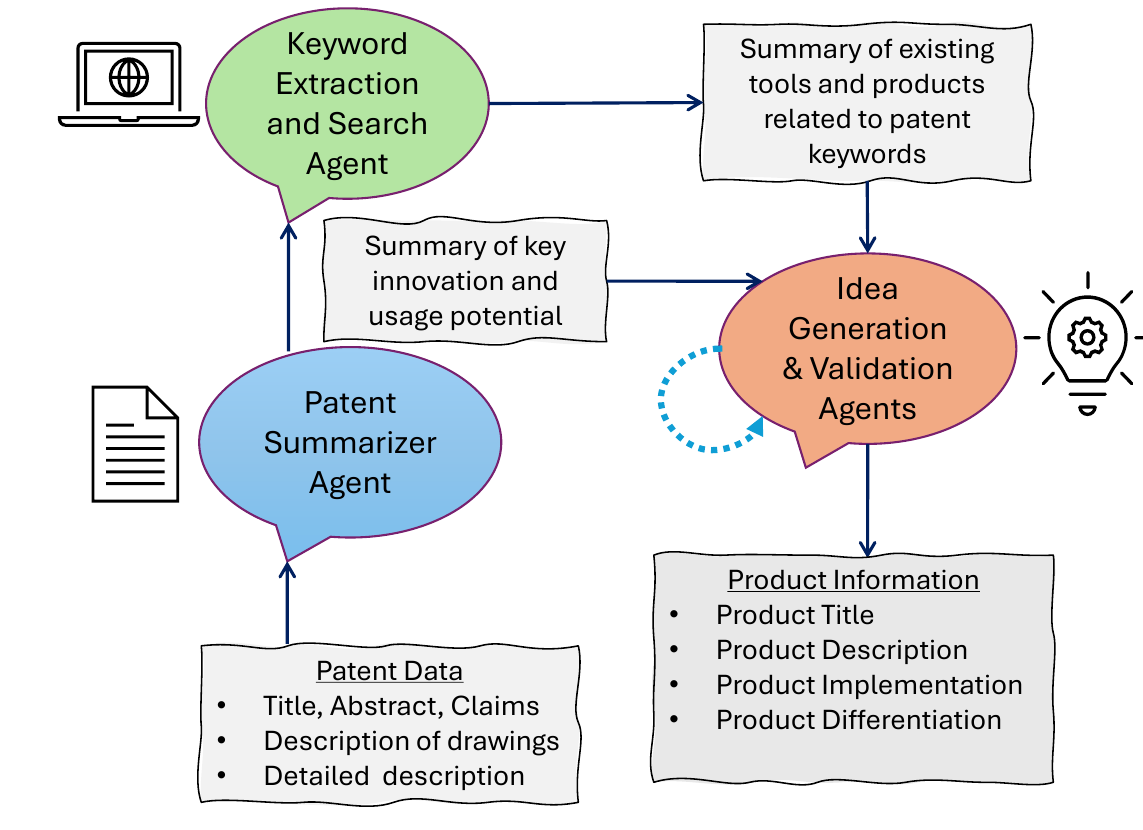}  
    \caption{Illustration of the Agent Ideate Pipeline.}  
    \label{fig:agent_ideate}
    % \vspace{-5mm}
\end{figure}
To achieve this, AgentScen 2025 shared task\footnote{\url{https://sites.google.com/view/agentscen/shared-task}} on Product Business Idea Generation from Patents (PBIG) was introduced as part of the $2^{nd}$ Workshop on Agent AI for Scenario Planning at IJCAI-25. 
\subsection{Task formulation}
The goal of this task is to evaluate systems that can read a patent and generate a realistic product idea that could be implemented and launched within three years. Each submission is expected to produce four concise outputs for a given patent: 
\begin{enumerate}
    \item \textbf{\textit{Product title}}: A concise name for the product.
    \item \textbf{\textit{Product description}}: A brief explanation of the product outlining its essential features, target users, their needs, and the benefits provided by the product.
    \item \textbf{\textit{Implementation}}: An explanation describing the implementation of patents technology into the product.
    \item \textbf{\textit{Differentiation}}: An explanation highlighting what makes the product unique.
\end{enumerate}

To support this task, the organizers released a curated dataset consisting of 150 U.S. patents across three categories: Computer Science (CS), Natural Language Processing (NLP), and Material Chemistry. Participants were allowed to use external resources to enhance idea generation. System outputs were evaluated by both human experts and LLM-based evaluators based on multiple criteria, including technical feasibility, innovation, specificity, market need, and competitive advantage.
% \begin{enumerate}
%     \item Framework Diagram
%     \item Gen AI capabilities
%     \item The task of idea generation from patents/papers
%     \item Challenges involved (trustworthy aspects?)
%     \item Existing methods
% \end{enumerate}

% This paper presents our system developed for the PBIG shared task as the ``\textbf{TrustAI}" team. We have experimented with various approaches including prompt-based LLM, Multi-Agent architecture, and Multi-Agent architecture leveraging external search tool for generating the product ideas from patent text. The pipeline diagram is presented in Figure \ref{fig:agent_ideate}. We leverage LLM as a judge approach to evaluate the ideas generated by various approaches to select the best approach. In the following sections we discuss existing works, dataset, and details of our experimental setup and discussion of results. and analyzes the effectiveness of agent-based and LLM-driven architectures for transforming patent knowledge into innovative product concepts.
% This paper presents our system developed for the PBIG shared task as the ``\textbf{TrustAI}'' team. 
In this study, we built the \textbf{Agent Ideate} framework, which is a Multi-Agent architecture leveraging an external search tool for generating product ideas from patent text. The pipeline diagram is illustrated in Figure \ref{fig:agent_ideate}. We leverage an LLM-based judging approach to evaluate the ideas generated by the different methods and to select the most effective one. 
% In the following sections, we discuss related work, the dataset, details of our experimental setup, and a discussion of the results. 
We also analyze the effectiveness of agent-based and LLM-driven architectures for transforming patent knowledge into innovative product concepts. Our code is publicly available \footnote{\url{https://github.com/gopichandkanumolu/AgentIdeate}}.

\section{Related Work}
The task of generating business ideas from patent documents \cite{yoshiyasu2025nsnlp, xu2025mk2, terao2025collaborative, hoshino2025multiagent, shimanuki2025selfimprovement} intersects with multiple research domains, including patent analysis \cite{sheremetyeva-2003-natural}, patent summarization \cite{sharma-etal-2019-bigpatent}, knowledge extraction \cite{tonguz-etal-2021-automating}, and large language model (LLM)-driven ideation. Prior work has explored the use of NLP and information retrieval techniques to extract technical concepts \cite{suzuki-takatsuka-2016-extraction, tonguz-etal-2021-automating} and commercial potential applications from patent texts \citep{1.souili2015natural, jiang-etal-2025-large}. More recently, LLMs have been applied for creative tasks such as product ideation, innovation support, showing promise in structured content generation \citep{2.girotra2023ideas, 3.radensky2024scideator, 4.li2024chain, wen2006generating}.

% However, few studies focus specifically on the end-to-end pipeline of transforming dense patent text into structured, market-aligned business ideas. Our work builds on these foundations by evaluating both prompt-based and agent-based LLM approaches, and by incorporating external tool usage such as web search to enhance contextual grounding and novelty of the generated ideas.

One closely related line of work is by \citet{5.si2024can}, who investigate the research ideation capabilities of LLMs. They pose a critical question: Are current LLMs capable of generating novel ideas that rival those produced by human experts? To answer this, the authors conducted a large-scale study involving over 100 qualified NLP researchers who generated human baselines and performed blind evaluations of both human and LLM-generated ideas. Their findings reveal that LLM-generated ideas are often judged as more novel than those produced by domain experts. 
% The study introduces a research ideation agent with three key components: paper retrieval, idea generation, and idea ranking, and demonstrates that LLMs can achieve competitive or superior performance to humans across most evaluation dimensions.

SciMON\citep{wang-etal-2024-scimon} is a framework that enhances language models' ability to generate novel scientific ideas by leveraging literature-based inspirations and iterative novelty optimization. Unlike traditional link-prediction approaches, it takes contextual inputs (e.g., research problems) and produces natural language hypotheses, using retrieval from semantic, knowledge graph, and citation sources. While evaluations show improvements over GPT-4, the generated ideas still lack the depth and novelty of human-authored research. To this end, in contrast to the existing works, this study aims to generate product-based business ideas
from patents by building a multi-agentic framework.

% This work advances automated hypothesis generation but highlights remaining challenges in achieving scientific-level innovation.

% \begin{enumerate}
%     % \item A brief literature review
%     \item Summarize relevant research papers
% \end{enumerate}

\section{Dataset}
The dataset provided by the shared task organizers comprises a total of 150 U.S. patents, with 50 patents each from three distinct domains: Computer Science(CS), Natural Language Processing (NLP), and Material Chemistry(MC). Each patent entry includes structured metadata such as the title, abstract, claims, description, publication number, and publication date.

\noindent \textbf{Preprocessing}: Among these fields, the description section is notably extensive, often exceeding the input length limitations of most large language models (LLMs). To address this challenge, we implemented a preprocessing strategy that segments the description into semantically meaningful subsections. This was achieved through regular expression-based matching, which identifies and extracts parts such as: Background information, Brief description of drawings and claims, and Detailed description of the patent technology.

This segmentation allows for more efficient and focused processing by LLMs and downstream agents. Detailed statistics about the dataset distribution and content lengths across categories are summarized in Table \ref{tab:data-stats}.
\begin{table}[t]
\centering\small
\begin{tabular}{lrrl}
\toprule
\textbf{Section}       & \textbf{CS}  & \textbf{NLP}  & \textbf{Chemistry} \\ \midrule
Title                  &      10                    & 11            & 8                    \\ 
Abstract               &      134                   & 138           & 130                   \\ 
Background             &      1058                  & 910           & 6215                   \\ 
Claims                 &      1499                  & 1708          & 535                   \\ 
Description of Figures &      4636                  & 868           & 700                   \\ 
Detailed Description   &      1499                  & 5068          & 156                   \\ \bottomrule
\end{tabular}
\caption{Average number of words present in each section for different datasets. CS - Computer science, NLP - Natural Language Processing.}
\label{tab:data-stats}
\vspace{-5mm}
\end{table}

% \begin{enumerate}
%     \item Dataset Description
%     \item Preprocessing
%     \item Stats and Observations
% \end{enumerate}

\section{Methodology}
As presented in Figure \ref{fig:agent_ideate}, we adopt three distinct methods to generate innovative business ideas from patents. These methods are increasingly sophisticated in terms of architecture and capability:

\noindent \textbf{1. Prompt-based LLM Approach}: This is the simplest baseline. We use a single-prompt approach with a large language model (LLM), wherein the entire patent (or its reduced components: title, abstract, claims, and summarized description) is passed as input to the model. The prompt is crafted to guide the model in generating business ideas, specifying the required structure in JSON format with fields such as product title, product description, implementation, and differentiation.
    
\noindent \textbf{2. Multi-Agent LLM Architecture}: The second approach builds on modularization via a multi-agent system, where different tasks are handled by different specialized agents. Specifically:
    \begin{itemize}
        \item A Patent Analyst Agent summarizes the core innovation and usage of the patent.
        \item A Business Idea Generator Agent uses the summarized insight to generate a structured business idea.
        \item A Business Validator Agent ensures the output adheres to format, character limits, and originality constraints.
    \end{itemize}
    Each agent uses the same LLM backend but is provided with a distinct goal and context. Tasks are executed sequentially with inter-agent context passing, allowing for better modularity, reliability, and control compared to single-shot prompting. In the rest of the paper, we refer to this method as the \textit{Agent without Tool} approach.
    
\noindent \textbf{3. Multi-Agent LLM with External Search Tool}: The third and most comprehensive method incorporates a search tool to enrich the reasoning process with external information. It extends the second approach by introducing:
    \begin{itemize}
        \item A Keyword Extractor Agent, which identifies two core keywords from the summarized patent content.
        \item A Research Agent, which performs a DuckDuckGo tool-based web search using these keywords to gather information about existing tools, libraries, or products in the domain.
        \item The Business Idea Generator Agent utilizes both the patent summary and external market insights to create a business idea that is clearly differentiated from known solutions. 
        \item Finally, the Business Validator Agent ensures the output is well-formed, concise, and novel.
    \end{itemize}

% This setup leverages tool-augmented agents with external search, increasing awareness of existing solutions and thereby enhancing the quality and distinctiveness of generated ideas. 
We provide the role, goal, backstory, tool usage, task description, and expected output instructions for each agent in Appendix Table~\ref{tab:agent_config} and Table~\ref{tab:task_flow}. In the rest of the paper, we refer to this method as the \textit{Agent with Tool} approach.

% \noindent \textbf{Model Configuration:} Due to resource constraints and the lack of access to proprietary APIs such as OpenAI, we opted to experiment with open-source LLMs hosted via the Groq API\footnote{\url{https://console.groq.com/docs/models}}. Specifically, we used the llama-4-scout-17b-16e-instruct\footnote{\url{https://console.groq.com/docs/model/meta-llama/llama-4-scout-17b-16e-instruct}}
%  model for response generation in both architectures: the prompt-based LLM model and each agent in the multi-agent setup. The model was configured with a temperature of 0.7 and a maximum token limit of 1000. All experiments were conducted using the free-tier access provided by Groq.  For all agentic framework experiments, we used the CrewAI\footnote{\url{https://www.crewai.com/}} framework to create agents and integrate with external search tools.

% \textbf{To be continued....}

% \begin{enumerate}
%     % \item Describe about the 3 methods that are used
%     % \item Using Prompt-based LLM baseline
%     % \item Using Multi Agent Framework
%     % \item Using Multi Agent Framework with tool
%     % \item Clearly mention the prompts/roles/goals/backstory/tools used
%     % \item Generate a crew flow diagram from CrewAI
%     \item Mention the open-source llms used for experiments
%     \item Mention the configuration settings of LLMs
% \end{enumerate}

\begin{table*}[h]
\centering
\begin{tabular}{cllcc}
\toprule
\multicolumn{1}{c}{\textbf{Domain}} & \textbf{Idea 1} & \textbf{Idea 2} & \textbf{Idea 1 Count(\%)} & \textbf{Idea 2 Count(\%)} \\ \midrule
\multirow{3}{*}{Computer Science}     & Prompt-based LLM             & \textbf{Agent without tool}           &       14                &  \textbf{86}                     \\ 
                                      % & LLM             & Agent with Tool &                     &                     \\ 
                                      & Agent without tool           & \textbf{Agent with Tool} &         14              &   \textbf{86}                    \\ \midrule
\multirow{3}{*}{NLP}                  & Prompt-based LLM             & \textbf{Agent without tool}           &   02                    &    \textbf{98}                   \\ 
                                      % & LLM             & Agent with Tool &                       &                       \\ 
                                      & \textbf{Agent without tool}           & Agent with Tool &      \textbf{88}                 &      12                 \\ \midrule
\multirow{3}{*}{Material Chemistry}   & Prompt-based LLM             & \textbf{Agent without tool}           &        08               &  \textbf{92}                     \\ 
                                      % & LLM             & Agent with Tool &        4               &  46                     \\ 
                                      & \textbf{Agent without tool}           & Agent with Tool &         \textbf{64}              &      38                 \\ \bottomrule
\end{tabular}
\caption{Evaluation of ideas generated using various approaches. We employ the LLM-as-a-Judge method to compare the ideas and report the percentage of ideas selected by the judge.}
\label{tab:eval_stats}
\end{table*}

% #################### Oraganizer Human Evaluation Results ################
\begin{table}[h]
\centering
\resizebox{\columnwidth}{!}{
\begin{tabular}{cccc}
\toprule
\textbf{Criteria}     & \textbf{Chemistry} & \textbf{CS} & \textbf{NLP} \\ \midrule
Tech Validity         & 1                           & 2                         & 3                                    \\ 
Specificity        & 3                           & 3                         & 3                                    \\ 
Need Validation         & 5                           & 2                         & 4                                    \\ 
Market Size           & 5                           & 1                         & 1                                    \\ 
Innovativeness        & 5                           & 3                         & 4                                    \\ 
Competitive Advantage & 2                           & 3                         & 3                                    \\ \bottomrule
\end{tabular}
}
\caption{Human evaluation results provided by the organizers. Each row represents the rank/position of our submission "\textbf{TrustAI}" for each domain based on the scores for each criteria.}
% \caption{Human evaluation results provided by the organizers. Each row represents the rank/position of our submission for each domain based on the scores for each criteria.}
\label{tab:human_eval_results}
\vspace{-5mm}
\end{table}
\section{Experiments and Evaluation}
We conduct experiments with prompt-based, agent with Tool and agent without Tool based approaches. For all experiments, we used the llama-4-scout-17b-16e-instruct\footnote{\url{https://console.groq.com/docs/model/meta-llama/llama-4-scout-17b-16e-instruct}}
 model for response generation in both architectures: the prompt-based LLM model and each agent in the multi-agent setup. 
Due to resource constraints and the lack of access to proprietary APIs such as OpenAI, we opted to experiment with open-source LLMs hosted via the Groq API\footnote{\url{https://console.groq.com/docs/models}}. The LLM was configured with a temperature of 0.7 and a maximum token limit of 1000. All experiments were conducted using the free-tier access provided by Groq.  For all agentic framework experiments, we used the CrewAI\footnote{\url{https://www.crewai.com/}} framework to create agents and integrate with external search tools.

To assess the relative quality of business ideas generated by different methods, we employed an LLM-as-a-judge evaluation strategy. Specifically, we designed a structured prompt where the model is provided with a patent description and two product ideas generated using different approaches (e.g., baseline prompting vs. multi-agent with search). The LLM is then instructed to critically evaluate the ideas across six well-defined dimensions: technical validity, innovativeness, specificity, need validity, market size, and competitive advantage.

% The evaluation prompt is carefully engineered to mimic the reasoning process of a domain expert. By explicitly listing the evaluation criteria, we reduce ambiguity and encourage the model to weigh each dimension before issuing a verdict. The output is expected in a strict JSON format, containing both the selected better idea (idea 1 or idea 2) and a rationale explaining the decision. 

% The evaluation setup and the description of the evaluation criteria are provided in Appendix Table~\ref{tab:eval_llm} and Table~\ref{tab:eval_criteria}. By explicitly listing the evaluation criteria, we reduce ambiguity and encourage the model to weigh each dimension before issuing a verdict. The output is expected in a strict JSON format, containing both the selected better idea (idea 1 or idea 2) and a rationale explaining the decision. 

The evaluation setup and criteria are provided in Appendix Table~\ref{tab:eval_llm} and Table~\ref{tab:eval_criteria}. Explicitly listing the criteria reduces ambiguity and encourages the model to weigh each dimension before issuing a verdict. The output follows a strict JSON format, containing the selected better idea (idea 1 or idea 2) and a rationale for the decision.

We used a high-capacity model LLaMA 3 70B \footnote{\url{https://console.groq.com/docs/model/llama-3.3-70b-versatile}} hosted via Groq for inference, ensuring strong reasoning and evaluation capabilities. This method of LLM-based comparative evaluation offers a scalable and cost-effective alternative to human annotation, especially in scenarios involving nuanced technical and entrepreneurial judgments. Furthermore, by leveraging LLMs that are blind to the origin of each idea, we minimize bias and ensure that comparisons focus purely on idea quality, not model provenance.

% \begin{table*}[t]
% \centering
% \begin{tabular}{l|c|c|c}
% \hline
% \textbf{Section}       & \textbf{LLM baseline Vs Agent}  & \textbf{LLM baseline Vs Agent with Tool}  & \textbf{Agent Vs Agent with Tool} \\ \hline
% idea 1                 &      c1                    & c1            & c1                   \\ \hline
% idea 2                 &      c2                  & c2          & c2                   \\ \hline
% \end{tabular}
% \caption{Evaluation of generated ideas using LLM as a judge.}
% \label{tab:eval_stats}
% \end{table*}

\section{Discussion}

\textbf{Evaluation using LLM as a judge}:
    The automated evaluation results (using LLM as judge) in Table \ref{tab:eval_stats} show clear performance differences between approaches. The Agent with Tool method consistently generates highly-ranked ideas in Computer Science (86\%), demonstrates moderate performance in Material Chemistry (38\%), but performs poorly in NLP (12\%). The standalone Agent approach without tool usage shows strong performance in NLP (98\%) and Material Chemistry (64\%), though it is less effective in Computer Science (14\%) compared to the Agent with Tool method. The basic LLM prompt method performs poorly across all domains (Computer Science: 14\%, NLP: 02\%, Material Chemistry: 08\%), suggesting that multi-agent frameworks provide substantial benefits even without tool access.

    Based on the automatic evaluation results comparing which approach generated the best ideas for each domain, we submitted the highest-performing outputs for organizer evaluation. The results of this evaluation are discussed in the following section.

\noindent \textbf{Evaluation results given by the Organizers}:
The human evaluation rankings in Table \ref{tab:human_eval_results} reveal important domain-specific patterns. In Chemistry, our system achieved top rankings in Innovativeness ($1^{st}$) but performed poorly in Technical Validity ($5^{th}$), indicating highly creative but potentially less feasible ideas. For Computer Science, we see balanced performance across criteria (mostly $2^{nd}$-$3^{rd}$ place), suggesting reliable but not exceptional results. The NLP domain shows our strongest overall performance, with top-3 rankings in all criteria except Market Size ($5^{th}$), highlighting both the technical strength and potential niche focus of generated ideas.

\section{Conclusion}
This paper presented our framework Agent Ideate, for generating product ideas from patents. We have conducted experiments using prompt-based LLM, multi-agent framework, and tool-augmented agents. Automated evaluation (LLM-as-judge) showed that Agent with Tool performed best in Computer Science, while standalone Agent excelled in NLP, and Material Chemistry. Our findings highlight the potential of agentic AI for structured innovation while underscoring domain-specific challenges.

\section{Limitations}
Our study has several key limitations. First, reliance on open-source LLMs (e.g., LLama-4-17B, and LLaMA-3-70B) may restrict performance compared to state-of-the-art proprietary models. Second, the system’s effectiveness varies significantly across domains, requiring domain specific models. Finally, the tool-augmented agent’s performance depends heavily on external search quality, which can introduce noise. These constraints highlight the need for more robust domain adaptation, hybrid evaluation methods, and improved tool integration in future work.

\bibliography{custom}

\begin{thebibliography}{20}
\providecommand{\natexlab}[1]{#1}

\bibitem[{Girotra et~al.(2023)Girotra, Meincke, Terwiesch, and Ulrich}]{2.girotra2023ideas}
Karan Girotra, Lennart Meincke, Christian Terwiesch, and Karl~T Ulrich. 2023.
\newblock \href {https://christophegirard.com/wp-content/uploads/2023/09/Etude-creation-idees-comparative-ChatGPT-vs-etudiants.pdf} {Ideas are dimes a dozen: Large language models for idea generation in innovation}.
\newblock \emph{The Wharton School Research Paper Forthcoming}.

\bibitem[{Hoshino et~al.(2025)Hoshino, Shramatsu, and Nagasawa}]{hoshino2025multiagent}
Mizuki Hoshino, Shun Shramatsu, and Fuminori Nagasawa. 2025.
\newblock A business idea generation framework based on creative multi-agent discussions.
\newblock In \emph{The 2nd Workshop on Agent AI For Scenario Planning (AGENTSCEN2025)}.

\bibitem[{Jiang and Goetz(2024)}]{jiang2024artificial}
Lekang Jiang and Stephan Goetz. 2024.
\newblock \href {https://arxiv.org/abs/2403.04105} {Artificial intelligence exploring the patent field}.
\newblock \emph{arXiv e-prints}, pages arXiv--2403.

\bibitem[{Jiang et~al.(2025)Jiang, Zhang, Scherz, and Goetz}]{jiang-etal-2025-large}
Lekang Jiang, Caiqi Zhang, Pascal~A. Scherz, and Stefan Goetz. 2025.
\newblock \href {https://doi.org/10.18653/v1/2025.findings-naacl.70} {Can large language models generate high-quality patent claims?}
\newblock In \emph{Findings of the Association for Computational Linguistics: NAACL 2025}, pages 1272--1287, Albuquerque, New Mexico. Association for Computational Linguistics.

\bibitem[{Li et~al.(2024)Li, Xu, Guo, Zhao, Li, Yuan, Zhang, Jiang, Xin, Dang et~al.}]{4.li2024chain}
Long Li, Weiwen Xu, Jiayan Guo, Ruochen Zhao, Xingxuan Li, Yuqian Yuan, Boqiang Zhang, Yuming Jiang, Yifei Xin, Ronghao Dang, and 1 others. 2024.
\newblock \href {https://arxiv.org/abs/2410.13185} {Chain of ideas: Revolutionizing research via novel idea development with llm agents}.
\newblock \emph{arXiv preprint arXiv:2410.13185}.

\bibitem[{Mossoff(2000)}]{mossoff2000rethinking}
Adam Mossoff. 2000.
\newblock \href {https://repository.uclawsf.edu/hastings_law_journal/vol52/iss6/2/} {Rethinking the development of patents: an intellectual history, 1550-1800}.
\newblock \emph{Hastings Lj}, 52:1255.

\bibitem[{Radensky et~al.(2024)Radensky, Shahid, Fok, Siangliulue, Hope, and Weld}]{3.radensky2024scideator}
Marissa Radensky, Simra Shahid, Raymond Fok, Pao Siangliulue, Tom Hope, and Daniel~S Weld. 2024.
\newblock \href {https://arxiv.org/abs/2409.14634} {Scideator: Human-llm scientific idea generation grounded in research-paper facet recombination}.
\newblock \emph{arXiv preprint arXiv:2409.14634}.

\bibitem[{Sharma et~al.(2019)Sharma, Li, and Wang}]{sharma-etal-2019-bigpatent}
Eva Sharma, Chen Li, and Lu~Wang. 2019.
\newblock \href {https://doi.org/10.18653/v1/P19-1212} {{BIGPATENT}: A large-scale dataset for abstractive and coherent summarization}.
\newblock In \emph{Proceedings of the 57th Annual Meeting of the Association for Computational Linguistics}, pages 2204--2213, Florence, Italy. Association for Computational Linguistics.

\bibitem[{Sheremetyeva(2003)}]{sheremetyeva-2003-natural}
Svetlana Sheremetyeva. 2003.
\newblock \href {https://doi.org/10.3115/1119303.1119311} {Natural language analysis of patent claims}.
\newblock In \emph{Proceedings of the {ACL}-2003 Workshop on Patent Corpus Processing}, pages 66--73, ,. Association for Computational Linguistics.

\bibitem[{Shimanuki et~al.(2025)Shimanuki, Shimizu, Kinugasa, and Sugisawa}]{shimanuki2025selfimprovement}
Masaya Shimanuki, Naoto Shimizu, Kentaro Kinugasa, and Hiroki Sugisawa. 2025.
\newblock Business idea generation from patent documents: Knowledge integration and self-improvement via llm.
\newblock In \emph{The 2nd Workshop on Agent AI For Scenario Planning (AGENTSCEN2025)}.

\bibitem[{Si et~al.(2024)Si, Yang, and Hashimoto}]{5.si2024can}
Chenglei Si, Diyi Yang, and Tatsunori Hashimoto. 2024.
\newblock \href {https://arxiv.org/abs/2409.04109} {Can llms generate novel research ideas? a large-scale human study with 100+ nlp researchers}.
\newblock \emph{arXiv preprint arXiv:2409.04109}.

\bibitem[{Souili et~al.(2015)Souili, Cavallucci, and Rousselot}]{1.souili2015natural}
Achille Souili, Denis Cavallucci, and Fran{\c{c}}ois Rousselot. 2015.
\newblock \href {https://www.sciencedirect.com/science/article/pii/S1877705815043490} {Natural language processing (nlp)--a solution for knowledge extraction from patent unstructured data}.
\newblock \emph{Procedia engineering}, 131:635--643.

\bibitem[{Suzuki and Takatsuka(2016)}]{suzuki-takatsuka-2016-extraction}
Shoko Suzuki and Hiromichi Takatsuka. 2016.
\newblock \href {https://aclanthology.org/C16-1113/} {Extraction of keywords of novelties from patent claims}.
\newblock In \emph{Proceedings of {COLING} 2016, the 26th International Conference on Computational Linguistics: Technical Papers}, pages 1192--1200, Osaka, Japan. The COLING 2016 Organizing Committee.

\bibitem[{Terao and Tachioka(2025)}]{terao2025collaborative}
Yasunori Terao and Yuuki Tachioka. 2025.
\newblock Collaborative invention: Refining patent-based product ideation via llm-guided selection and rewriting.
\newblock In \emph{The 2nd Workshop on Agent AI For Scenario Planning (AGENTSCEN2025)}.

\bibitem[{Tonguz et~al.(2021)Tonguz, Qin, Gu, and Moon}]{tonguz-etal-2021-automating}
Ozan Tonguz, Yiwei Qin, Yimeng Gu, and Hyun~Hannah Moon. 2021.
\newblock \href {https://doi.org/10.18653/v1/2021.nllp-1.21} {Automating claim construction in patent applications: The {CMU}mine dataset}.
\newblock In \emph{Proceedings of the Natural Legal Language Processing Workshop 2021}, pages 205--209, Punta Cana, Dominican Republic. Association for Computational Linguistics.

\bibitem[{Urlana et~al.(2024)Urlana, Kumar, Singh, Garlapati, Chalamala, and Mishra}]{urlana2024llms}
Ashok Urlana, Charaka~Vinayak Kumar, Ajeet~Kumar Singh, Bala~Mallikarjunarao Garlapati, Srinivasa~Rao Chalamala, and Rahul Mishra. 2024.
\newblock \href {https://arxiv.org/abs/2402.14558} {Llms with industrial lens: Deciphering the challenges and prospects--a survey}.
\newblock \emph{arXiv preprint arXiv:2402.14558}.

\bibitem[{Wang et~al.(2024)Wang, Downey, Ji, and Hope}]{wang-etal-2024-scimon}
Qingyun Wang, Doug Downey, Heng Ji, and Tom Hope. 2024.
\newblock \href {https://doi.org/10.18653/v1/2024.acl-long.18} {{S}ci{MON}: Scientific inspiration machines optimized for novelty}.
\newblock In \emph{Proceedings of the 62nd Annual Meeting of the Association for Computational Linguistics (Volume 1: Long Papers)}, pages 279--299, Bangkok, Thailand. Association for Computational Linguistics.

\bibitem[{Wen et~al.(2006)Wen, Jiang, Wen, and Shadbolt}]{wen2006generating}
Guihua Wen, Lijun Jiang, Jun Wen, and Nigel~R Shadbolt. 2006.
\newblock \href {https://dl.acm.org/doi/abs/10.5555/1757898.1757978} {Generating creative ideas through patents}.
\newblock In \emph{Proceedings of the 9th Pacific Rim international conference on Artificial intelligence}, pages 681--690.

\bibitem[{Xu et~al.(2025)Xu, Hirasawa, Kawano, Kato, and Kozuno}]{xu2025mk2}
Yuzheng Xu, Tosho Hirasawa, Seiya Kawano, Shota Kato, and Tadashi Kozuno. 2025.
\newblock Mk2 at pbig competition: A prompt generation solution.
\newblock In \emph{The 2nd Workshop on Agent AI For Scenario Planning (AGENTSCEN2025)}.

\bibitem[{Yoshiyasu(2025)}]{yoshiyasu2025nsnlp}
Hayato Yoshiyasu. 2025.
\newblock Team ns\_nlp at the agentscen shared task: Structured ideation using divergent and convergent thinking.
\newblock In \emph{The 2nd Workshop on Agent AI For Scenario Planning (AGENTSCEN2025)}.

\end{thebibliography}

\appendix

\section{Appendix}
\label{sec:appendix}
We present the description of each agent’s role, goal, backstory, and the tools they can access in Table~\ref{tab:agent_config} and Table~\ref{tab:task_flow}. This also includes the task descriptions and expected outputs for each agent. Additionally, we provide the evaluation criteria used to compare the ideas generated by various methods using the LLM-as-a-judge approach in Table~\ref{tab:eval_llm} and Table~\ref{tab:eval_criteria}.

\begin{table*}[htbp]
\centering
\renewcommand{\arraystretch}{1.3}
\begin{tabularx}{\textwidth}{@{}>{\raggedright\arraybackslash}p{2.5cm} >{\raggedright\arraybackslash}p{2.5cm} X X@{}}
\rowcolor{headergray}
\textbf{Agent Name} & \textbf{Role} & \textbf{Goal} & \textbf{Backstory / Tools Used} \\
\midrule
\rowcolor{row1}
Patent Analyst & Reader Agent & Extract and summarize key features from patents & Specializes in understanding complex patent documents and identifying key technological aspects. \\
\rowcolor{row2}
Keyword Extractor & Keyword Agent & Generate essential keywords from patent summary & NLP expert identifying core technologies to support product discovery. \\
\rowcolor{row3}
Researcher & Search Agent & Search for relevant products/tools using keywords and synthesize results & Enthusiast in discovering tools/products relevant to keywords with clear and concise summaries. Tools Used: DuckDuckGo Tool \\
\rowcolor{row4}
Idea Generator & Business Idea Agent & Generate innovative product ideas from patent content & Creative entrepreneur skilled in mapping technology to business ideas.\\
\rowcolor{row5}
Business Validator & Validator Agent & Validate ideas for structure and uniqueness & Ensures business ideas are well-formatted, feasible, and differentiated from existing solutions.\\
\bottomrule
\end{tabularx}
% \caption{Agent Configuration}
\caption{Description of each agent’s role, goal, backstory, and tool usage}
\label{tab:agent_config}
\end{table*}

\begin{table*}[htbp]
\centering
\renewcommand{\arraystretch}{1.3}
\begin{tabularx}{\textwidth}{@{}>{\raggedright\arraybackslash}p{3cm} >{\raggedright\arraybackslash}p{3cm} X X@{}}
\rowcolor{headergray}
\textbf{Task Name} & \textbf{Performed By} & \textbf{Task Description} & \textbf{Expected Output} \\
\midrule
\rowcolor{row1}
Patent Analysis & Patent Analyst & Read and extract core information from patent sections & Structured summary of key patent features.\\
\rowcolor{row2}
Keyword Generation & Keyword Extractor & Generate two keywords representing the patent’s core technological concepts & List of keywords: \texttt{["keyword1", "keyword2"]}\\
\rowcolor{row3}
Product Research & Researcher & Use keywords to search web using DuckDuckGo Tool for related products and synthesize findings & Text summary with relevant products/tools and short descriptions. \\
\rowcolor{row4}
Idea Generation & Idea Generator & Based on findings and patent, generate an innovative product/business idea & JSON object with below fields: \texttt{product\_title}, \texttt{product\_description}, \texttt{implementation}, \texttt{differentiation}.\\
\rowcolor{row5}
Idea Validation & Business Validator & Review the generated idea for adherence to format and uniqueness & Validated JSON output with feedback on issues if any.\\
\bottomrule
\end{tabularx}
% \caption{Task Flow Configuration}
\caption{Description of each agent’s task, and expected output for each task.}
\label{tab:task_flow}
\end{table*}

\begin{table*}[t]
\centering
\renewcommand{\arraystretch}{1.3}
\begin{tabularx}{\textwidth}{@{}>{\raggedright\arraybackslash}p{4cm} X@{}}
\rowcolor{headergray}
\textbf{Aspect} & \textbf{Description} \\
\midrule
\rowcolor{row1}
Evaluator Role & LLM-as-a-Judge: A large language model is prompted to objectively compare two product ideas derived from a common patent. \\
\rowcolor{row2}
Input Provided & 1. Patent description \newline 2. Two distinct product/business ideas using the patent \\
\rowcolor{row3}
Evaluation Goal & Select the better idea based on well-defined business and technical criteria. \\
\rowcolor{row4}
Prompt Structure & Multi-section prompt including: \newline • \texttt{<patent>}: full patent description \newline • \texttt{<idea\_1>}, \texttt{<idea\_2>}: structured product ideas \newline • Explicit list of 6 evaluation criteria (refer Table~\ref{tab:eval_criteria}) \\
\rowcolor{row5}
LLM Output Format & JSON: \texttt{\{"output": "idea\_1 or idea\_2", "reason": "reason for the choice"\}} \\
\rowcolor{row1}
Use Case & Used for comparative evaluation of generated product ideas, testing how well different agents or models transform patent knowledge into viable business ideas. \\
\bottomrule
\end{tabularx}
\caption{Evaluation (LLM-as-a-Judge) Setup Overview}

\label{tab:eval_llm}
\end{table*}

\begin{table*}[htbp]
\centering
\renewcommand{\arraystretch}{1.3}
\begin{tabularx}{\textwidth}{@{}>{\raggedright\arraybackslash}p{4cm} X@{}}
\rowcolor{headergray}
\textbf{Criterion} & \textbf{Explanation} \\
\midrule
\rowcolor{row1}
Technical Validity & Is the patent technology appropriate and realistically implementable within 3 years? \\
\rowcolor{row2}
Innovativeness & Does the idea utilize the patent in a novel way? Does it stand out in terms of technological creativity? \\
\rowcolor{row3}
Specificity & Is the idea clearly and narrowly defined (e.g., “manage references” vs. “do research”)? \\
\rowcolor{row4}
Need Validity & Is there a clear and valid user need addressed by the product idea? \\
\rowcolor{row5}
Market Size & Is the target market large enough to make the product viable? Are there many potential users? \\
\rowcolor{row1}
Competitive Advantage & Does the use of the patented technology offer a unique advantage over competitors? \\
\bottomrule
\end{tabularx}
% \caption{Evaluation Criteria Breakdown}
\caption{Description of evaluation criteria of generated ideas using LLM as a judge.}
\label{tab:eval_criteria}
\end{table*}

\end{document}